\documentclass[runningheads]{llncs}
\usepackage{graphicx}
\usepackage{amsmath}
\usepackage{multirow}
\usepackage{tabularx}
\begin{document}
\title{Mention-centered Graph Neural Network for Document-level Relation Extraction}
\titlerunning{Mention-centered Graph Neural Network}
%
\author{Jiaxin Pan\inst{1} \and
Min Peng\inst{1} \and
Yiyan Zhang\inst{2}}
%
%
\institute{Computer School, Wuhan University \email{pjx\_1997, peng@whu.edu.cn}\and
National University of Singapore
\email{e0261914@u.nus.edu}}
\maketitle              
\begin{abstract}
Document-level relation extraction aims to discover relations between entities across a whole document. How to build the dependency of entities from different sentences in a document remains to be a great challenge. Current approaches either leverage syntactic trees to construct document-level graphs or aggregate inference information from different sentences. In this paper, we build cross-sentence dependencies by inferring compositional relations between inter-sentence mentions. Adopting aggressive linking strategy, intermediate relations are reasoned on the document-level graphs by mention convolution. We further notice the generalization problem of NA instances, which is caused by incomplete annotation and worsened by fully-connected mention pairs. An improved ranking loss is proposed to attend this problem. Experiments show the connections between different mentions are crucial to document-level relation extraction, which enables the model to extract more meaningful higher-level compositional relations.


\end{abstract}
\section{Introduction}
\begin{figure}
   \centering
    \includegraphics[width=1\textwidth]{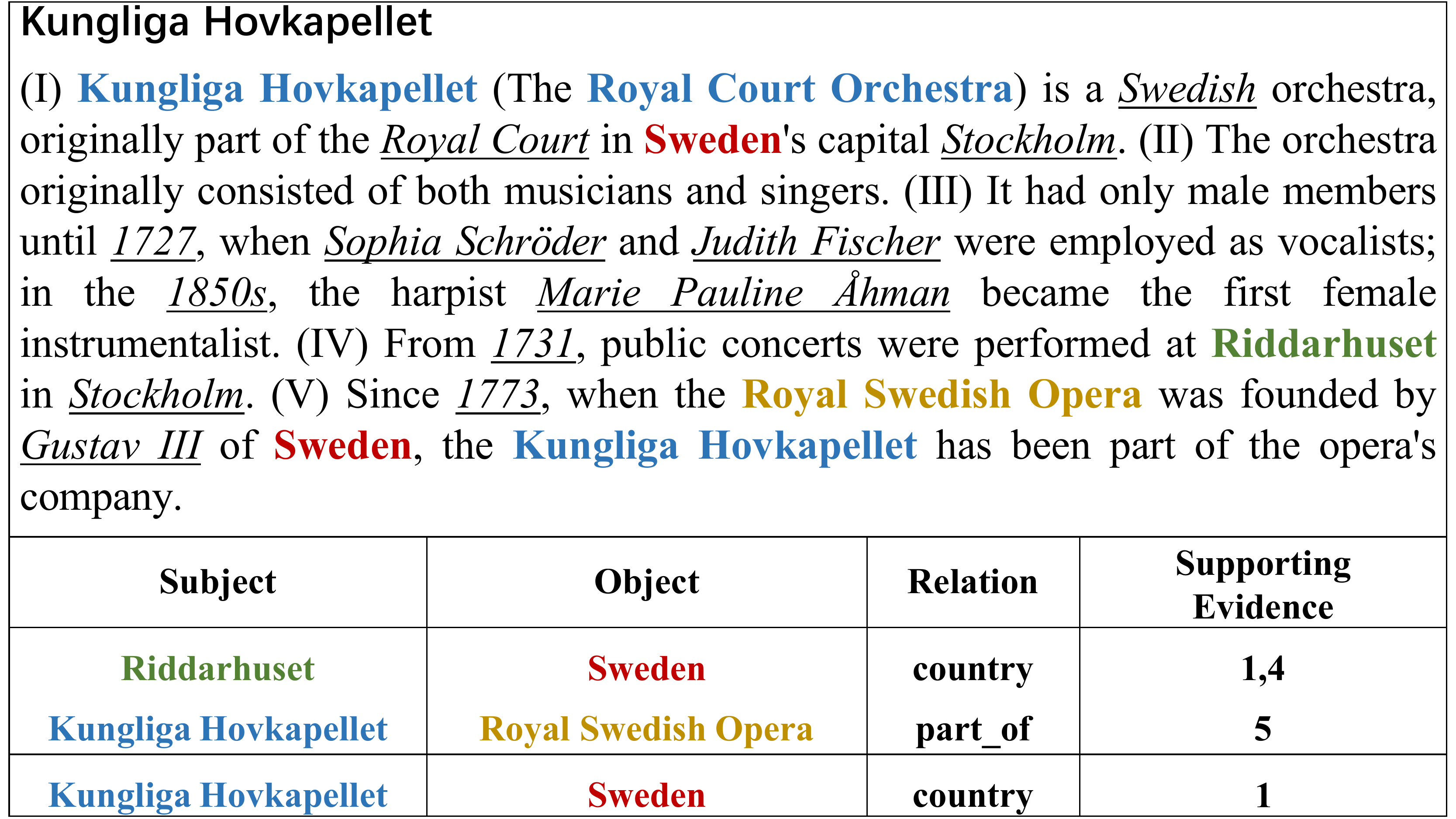}
    \caption{An example document from DocRED. The first 2 relational facts from annotations are presented with named entity mentions involved in various colors. Other named entity mentions are underlined. The last instance is not included in the original annotation. Supporting evidences of each relation are provided as well.}
    \label{fig:docred_example}
\end{figure}

Relation Extraction(RE) is the task of predicting relations between named entities in plain text. It is an important task for many downstream applications such as knowledge graph construction \cite{luan-etal-2018-multi}
and question answering \cite{yu-etal-2017-improved}. Most existing approaches  \cite{zeng-etal-2014-relation,zhou-etal-2016-attention,lin-etal-2016-neural,phi-etal-2018-ranking} focus on sentence-level RE, which discover relational facts from a single sentence. However, as is shown in Figure 
 \ref{fig:docred_example}, in real world scenario, many relational facts lie in different sentences in a document. The task of identifying such relations is called document-level RE. To accelerate the development of document-level RE task, \cite{yao-etal-2019-docred} published the first and  so far unique dataset for large-scale document-level relation extraction, DocRED$\footnote{\url{https://github.com/thunlp/DocRED}}$, constructed from Wikipedia. 

Current baseline\cite{tang2020hin} on document-level relation extraction calculates attention scores between entity pairs and sentences to aggregate information through the whole document. Other efforts \cite{peng-etal-2017-cross,zheng2018effective} link the dependency trees of adjacent sentences to capture interactions among inter-sentence entities. However, we figure out the relations between inter-sentence  entities can be inferred directly from the intermediate relations between their coreference mentions. Taking Fig \ref{fig:docred_example} as an example, to infer the ``country'' relation between ``Riddarhuset'' and ``Sweden'', first we need to discover the relation ``in'' between ``Riddarhuset'' and ``Stockholm'' in sentence (\uppercase\expandafter{\romannumeral4}), the relation ``capital'' between ``Sweden'' and ``Stockholm'' in sentence(\uppercase\expandafter{\romannumeral1}) and finally make decisions through these intermediate relations. Unlike prior methods which explore their relations roundaboutly by dependency trees\cite{nan-etal-2020-reasoning} or aggregate sentence representation, we construct document-level graphs which connect inter-sentence mentions directly as cross-sentence dependencies and perform relational reasoning on the combination of mention nodes' representation. By transferring useful information step by step through connected mentions, compound relations can be spread to entity pairs' representations incrementally. We argue that document-level graph enables the model capture long-distance relational facts transcending adjacent limits and alleviates the noises of unrelated text.

We also notice the relational facts annotated by document-level dataset can hardly be complete due to the large number of potential entity pairs. For example, in DocRED, every document has an average of 26.2 entities, requiring 660.24 cross-sentence annotations. To alleviate the laborious work, human annotators are provided with recommendations from RE models and distant supervision based on entity linking. As a result, many NA instances in DocRED indeed have relations but they are not recommended by RE models or entity linking. Fig \ref{fig:docred_example} shows an example where (Kungliga  Hovkapellet, country, Sweden) is not included in original annotations. This means forcing the prediction score of mislabelling NA instances to 0 by traditional BCE loss may hurt the model's ability to generalize relational representations. The proposed fully-connected mention-centered document-level graph exacerbates the situation as it includes countless direct connections between mislabelling NA entity pairs and subordinate none NA mention pairs. To mitigate the problem, we design a new training objective which changes the classification task to ranking task, allowing the model to reach a balance between capturing the distribution of original annotations and preserving genuine relational representations.

Our contributions can be summarised as follows:
\begin{itemize}\setlength{\itemsep}{0pt}
    \item We propose a novel mention-centered model for document-level RE using fully-connected mention pairs to capture cross-sentence dependencies. By exchanging information between mentions pairs iteratively, entity pairs are capable of discovering more accurate inter-sentence relations. Our proposed model is independent of syntactic dependency tools and can achieve state-of-the-art performance on DocRED. We demonstrate the connections between mentions are the core component of inter-sentence relation extraction.
    \item We show detailed analysis about the incomplete annotation problem in DocRED, which interferes the generalization of NA instances. To relieve the negative impact of aggressive linking strategy on this problem, an improved version of ranking loss is proposed. Qualitative comparison between ranking loss and BCE loss further reveals the significance of the proposed training objective on our model.
\end{itemize}

\begin{figure}
    \centering
    \includegraphics[width=1\textwidth]{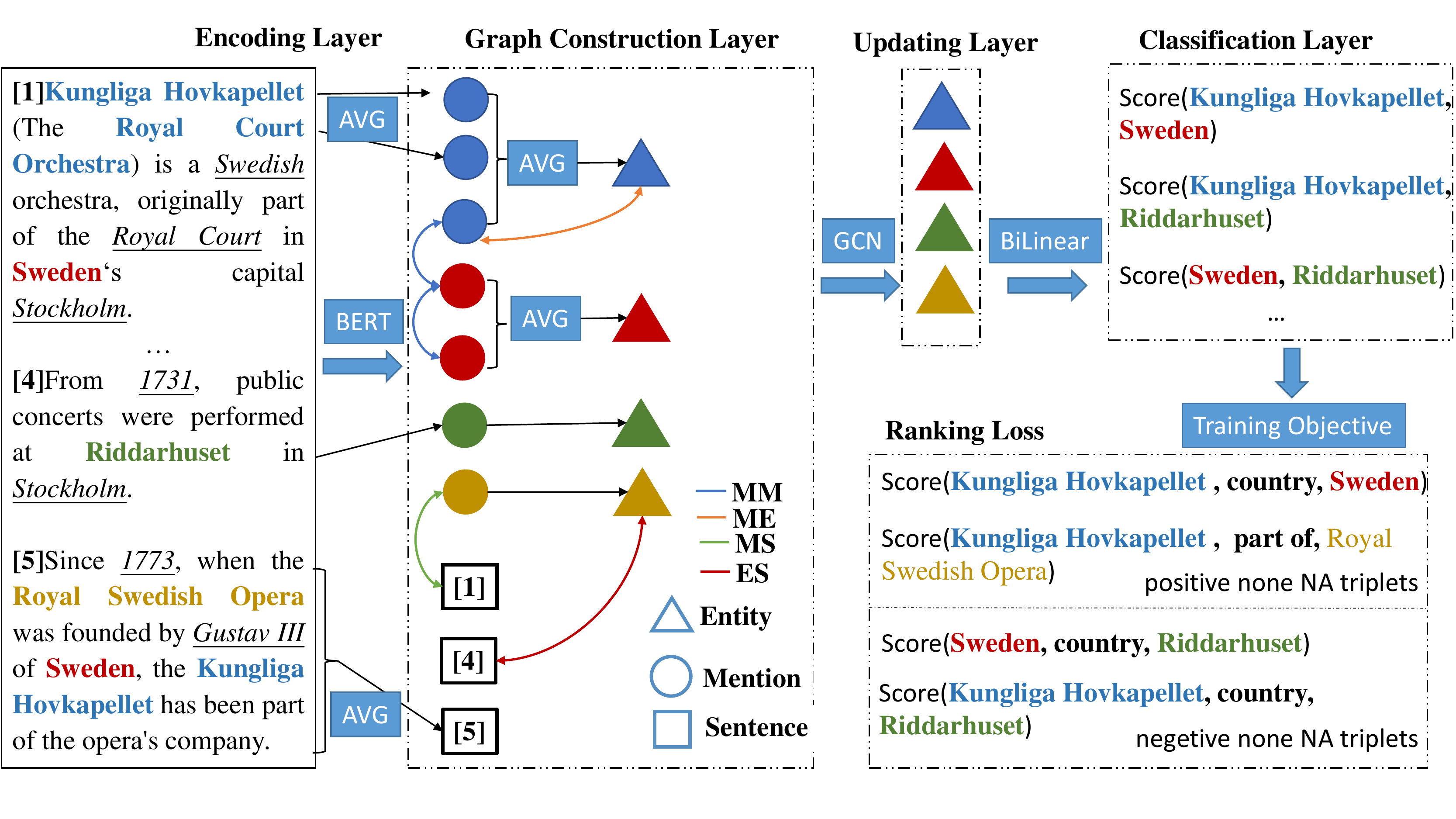}
    \caption{Overall architecture of the proposed model. The model first encodes the whole document by BERT. Then a document-level graph is constructed and fed into the updating layer to exchange information between different nodes. Finally, the classification layer calculates the probabilities of all entity pairs. Some node connections and self-loops are not shown for brevity.}
    \label{fig:model}
\end{figure}

\section{Model}
The overall architecture of proposed model is illustrated in Fig \ref{fig:model}. The proposed model consists of four layers: Encoding Layer, Graph Construction Layer, Updating Layer and Classification Layer. 

\subsection{Encoding Layer}
 Let $[x_{1}, x_{2}, ..., x_{n}]$ denote the document input, we use BERT to encode the document. A Linear layer is used to compress BERT embedding into low-dimensional space for afterwards processing. Entity type embeddings and coreference embeddings used in \cite{yao-etal-2019-docred} are concatenated afterwards. 

\begin{equation}
[h_{i}] = [W_{b}(\text{BERT}(x_{i})); e(i); c(i)]
\end{equation}
where $W_{b}$ is a trainable weight of linear map, $e$ and $c$ are embedding matrix for entity type and coreference type respectively. $[ ; ]$ denotes concatenation.

\subsection{Graph Construction Layer}
\subsubsection{Node Construction}
 We form three types of nodes in the graph: mention nodes($M$) $n_{m}$, entity nodes($E$) $n_{e}$, and sentence nodes($S$) $n_{s}$. A mention node $n_{m}$ ranging from the $s$-th word to the $t$-th word is represented as the average of hidden state from $s$ to $t$. The representation of an entity node $n_{e}$ is computed as the average of the mention representations associated with the entity. Finally, a sentence node $n_{s}$ is represented as the average of the word representations in the sentence. In order to distinguish different node types in the graph, we concatenate a node type $t$ embedding to each node representation. The final node representations are then estimated as $n_{m} = [avg_{w_{i}\in m}(w_{i}); t_{m}]$, $n_{e} = [avg_{w_{i}\in e}(w_{i}); t_{e}]$, $n_{s} = [avg_{m_{i}\in s}(w_{i}); t_{s}]$

\subsubsection{Edge Construction}
Accumulating compositional entity relations through representations of relevant mention pairs is the fundamental idea of our model. For this reason, we construct document-level graphs using the following 4 types of edges.

Mention-Mention($MM$):  To detect the implicit relations between mention pairs, we create mention-mention edges according to their relative distance. Unlike previous methods \cite{christopoulou-etal-2019-connecting,sahu-etal-2019-inter} which only connect mentions within a sentence or coreference mentions within an entity, our model creates mention-mention edges in an aggressive strategy by connecting every mention pairs in one document, as previous illustration about (Riddarhuset, country, Sweden) in Fig \ref{fig:docred_example} shows intermediate mentions can reside in different sentences or entities. In this way, document-level dependencies will be established through the chains of mentions which scattered in different sentences rather than sentence connection\cite{christopoulou-etal-2019-connecting} or the roots of parse trees between neighboring sentences\cite{sahu-etal-2019-inter}. We generate the edge representation between two mentions starting from $i$-th, $j$-th word as:

\begin{equation}
A_{M_{i}M_{j}}=\sigma\left(w_{\mathrm{m}}D(d_{ij}\right))
\end{equation}

where $w_{\mathrm{m}}$ are trainable model parameters and sigmoid activation function $\sigma$ is used, $d_{ij}$ are the relative distances of the mentions, $D$ is distance embedding matrix. In this way, $A_{i j}$ will be assigned to a real value between 0 and 1 according to their relative distance.  

Mention-Entity($ME$): To enable entities collect information gathered by subordinate mentions, we add ME edge. The edge between mention $i$ and entity $j$ is represented by:
{\setlength\abovedisplayskip{1pt plus 3pt minus 5pt}
\setlength\belowdisplayskip{1pt plus 3pt minus 5pt}}
\begin{equation}
    A_{M_{i}E_{j}} = \left\{
    \begin{array}{lr}
    1 ,&  i \in j \\
    0 ,&  i \notin j
    \end{array}
    \right.
\end{equation}

Mention-Sentence($MS$): If a mention belongs to a sentence, we intuitively think the sentence's representation encode the mention's relation information. We set the edge representation of mention $i$ and sentence $j$ as:
{\setlength\abovedisplayskip{1pt plus 3pt minus 5pt}
\setlength\belowdisplayskip{1pt plus 3pt minus 5pt}}
\begin{equation}
    A_{M_{i}S_{j}} = \left\{
    \begin{array}{lr}
    1 ,&  i \in j \\
    0 ,&  i \notin j
    \end{array}
    \right.
\end{equation}

Entity-Sentence($ES$): In the experiment we find that connecting entity nodes with their residing sentence nodes will improve the performance marginally. So, we set ES edge represented by:
{\setlength\abovedisplayskip{1pt plus 3pt minus 5pt}
\setlength\belowdisplayskip{1pt plus 3pt minus 5pt}}
\begin{equation}
    A_{E_{i}S_{j}} = \left\{
    \begin{array}{lr}
    1 ,&  i \in j \\
    0 ,&  i \notin j
    \end{array}
    \right.
\end{equation}

\subsection{Updating Layer}
To update the representations of entity pairs, we apply GCN operation on the constructed document-level graph. Vanilla GCN is designed for node classification task and weighs the importance of original nodes and neighbouring nodes equally. However, our interest is accumulating supplementary information to mention/entity nodes without losing their local expressive power. Besides, since $A_{ij}\leq 1$ in our model, the node representations will become smaller as the layers deepen especially when the node connects faraway mentions. To handle this problem, we integrate residual connections to original GCN operations:

\begin{equation}
h_{i}^{(k)} =ReLU 
            \left(\sum_{j=1}^{n} c_{i} A_{i j}\left(W^{(k)} h_{j}^{(k-1)}+b^{(k)}           \right) +h_{j}^{(k-1)}\right) 
\end{equation}
where $h_{j}^{(k)}$ is the embedding of node $j$ at the $k^{t h}$ layer, $b^{(k)}$ is a bias term, $W^{(k)}$ is a weight matrix. $c_{i}=1 / \sum_{j=1}^{n} A_{i j} $ is a normalization constant.

As pointed out in \cite{li2018deeper}, vanilla GCN operation makes the features of connected nodes similar. For one thing, we adopt this characteristic to synthesis higher-order information in the document-level graph. For another, by adding residual connections, we maintain original node representations rich of context information. 

\subsection{Classification Layer}
Following \cite{yao-etal-2019-docred}, we compute the probability of the given entity pair($e_{i}$, $e_{j}$) by sigmoid function. 
\begin{equation}
P(r|e_{i}, e_{j}) = sigmoid(N_{e_{i}}^{T} W_{t}N_{e_{j}} + b_{t})
\end{equation}
where $W_{t}$, $b_{t}$ are relation type dependent trainable weights and bias.

\subsection{Training Objective}
We divide the triplets produced by our model into 3 types: NA triplets, positive none NA triplets(the final outputs) and negative none NA triplets. In DocRED task setting, the output scores of none NA triplets are arranged in descending order. Triplets whose scores are greater than a threshold will be selected to the testing procedure and rest of candidates will be omitted as negative none NA class . Apparently, as long as the model ranks higher scores to annotations rather than  NA or negative none NA triplets, the results will be credible. Therefore, based on \cite{dos-santos-etal-2015-classifying}, we propose an improved version of ranking loss to simulate this procedure.

During a training batch $D$, $y^{+} \in D$ denotes the positive none NA triplets and  $y^{na} \in D$ denotes the negative none NA triplets. Let $s_{\theta}({y^{+}})$ and $s_{\theta}({y^{na}})$ be respective scores for triplets $y^{+}$ and $y^{na}$ generated by the network with parameter set $\theta .$ The new loss function is as follows:
\begin{equation}
L= \log \left(1+\exp\left(m^{+}-s_{\theta}({y^{+}})\right)\right) +\log \left(1+\exp\left(m^{-}+s_{\theta}({y^{na}})\right)\right) 
\end{equation}

where $m^{+}$ and $m^{-}$ are margins which help to measure the errors between predicted scores and labels. Training by minimizing this loss function will restrict the scores of negative none NA triplets smaller than those of positive none NA triplets. Scores of NA triplets are neglected because they will not be tested. In this way, we circumvent to make hard decisions of mislabelling NA triplets.

\section{Experiment}
In this section, we will introduce the DocRED dataset and model settings of our experiments.

\subsection{DocRED Dataset}
We use the DocRED dataset to evaluate the proposed method. DocRED contains 3,053 /1,000 /1,000 documents for training, development and test respectively, with 132,375 entities and 96 relation types.  Manual analysis shows about
40.7\% of relational facts can only be extracted from multiple sentences and 61.1\% relational instances require a variety of reasoning.

\subsection{Baselines}
\begin{itemize}\setlength{\itemsep}{0pt}
    \item CNN/LSTM/BiLSTM/Context-Aware\cite{yao-etal-2019-docred}: The CNN/LSTM/BiLSTM based models encode the whole document word by word with CNN/LSTM/BiLSTM as encoder. Context-Aware model\cite{sorokin-gurevych-2017-context} considers other relations in the context when predicting the target relation.  
    \item EoG\cite{christopoulou-etal-2019-connecting}/GCNN\cite{sahu-etal-2019-inter}: EoG connects sentence nodes in the document as inter-sentence dependencies and aggregates information through attention mechanism. GCNN constructs inter-sentence interactions by linking the roots of adjacent sentences’ parse trees and co-reference mentions within an entity, and then updates information through GCN.
    \item BERT/BERT-2step\cite{wang2019fine}:  \cite{wang2019fine} replaces the encoder with BERT\cite{devlin-etal-2019-bert}. BERT-2step further predicts if the relation exists between entity pairs before decides the accurate relation type of entity pairs.
    \item HIN\cite{tang2020hin}: HIN uses a hierarchical inference method to aggregate the inference information from entity, sentence, document levels respectively by attention mechanism and translation constraint.
    \item LSR\cite{nan-etal-2020-reasoning}: LSR treats the document-level graph structure as a latent variable and induces it through structured
    attention of shortest dependency path.
\end{itemize}

\subsection{Model Settings}
We use ``BERT-Base, Uncased'' version as BERT encoder in our experiments. The learning rate of BERT layer is $10^{-5}$ while the learning rate of GCN layer is $10^{-3}$. The embedding size of BERT model is 768. The layer number of GCN is set to be 2. We set $m^{+}$ to -1 and $m^{-}$ to -2. Other settings are the same as \cite{yao-etal-2019-docred}. ``+wiki'' means we use relation data from Wikidata $\footnote{\url{https://www.wikidata.org}}$ to facilitate the learning of ranking loss. 

\begin{table}
\caption{ Performance of different RE models on DocRED(\%). In Ignore F1 setting, relational facts appeared in training set are discarded during evaluation.}
\label{F1 score}
\centering
\begin{tabular}{|l|p{1.8cm}<{\centering}|p{1.8cm}<{\centering}|p{1.8cm}<{\centering}|p{1.8cm}<{\centering}|}
\hline
\multirow{2}{*}{\textbf{Model}}&\multicolumn{2}{c|}{\textbf{Dev}}& \multicolumn{2}{c|}{\textbf{Test}}\\
\cline{2-5}
  &\textbf{Ignore F1} & \textbf{F1} &\textbf{Ignore F1} & \textbf{F1} \\ 
\hline \hline
CNN\cite{yao-etal-2019-docred}&37.99  & 43.45& 36.44& 42.33\\
LSTM\cite{yao-etal-2019-docred}& 44.41& 50.66& 43.60& 50.12\\
BiLSTM\cite{yao-etal-2019-docred}& 45.12& 50.95& 44.73& 51.06\\
Context-Aware\cite{yao-etal-2019-docred}& 44.84& 51.10& 43.93& 50.64\\
\hline
EoG\cite{christopoulou-etal-2019-connecting}& 45.94 & 52.15 & 49.48 & 51.82\\
GCNN\cite{sahu-etal-2019-inter}& 46.00& 51.32& 49.79& 51.52\\
BiLSTM-LSR\cite{nan-etal-2020-reasoning}& 48.82& 55.17 & 52.15& 54.18\\
\hline
BiLSTM-MCN&51.89&54.00&51.24&53.54\\
BiLSTM-MCN+wiki&54.33&56.31&52.95&54.83\\
\hline
BERT\cite{wang2019fine}&-&54.16&-&53.20\\
BERT-2step\cite{wang2019fine}&-&54.42&-&53.92\\
HIN\cite{tang2020hin}&54.29&56.31&53.70&55.60\\
BERT-LSR\cite{nan-etal-2020-reasoning}&52.43&59.00&56.97&59.05\\
\hline
BERT-MCN&56.11&57.76&56.00&58.26\\
    - \textit{ranking loss}&55.32&57.00&55.20&57.50\\
BERT-MCN+wiki&\textbf{57.33}&\textbf{60.20}&\textbf{57.00}&\textbf{59.40}\\
\hline
\end{tabular}
\end{table}

\section{Result}
\subsection{Model Performance}
We use the same evaluation metrics as \cite{yao-etal-2019-docred} and evaluations on the test set are reported from CodaLab. 

Table \ref{F1 score} shows the results of different models under supervised settings. From the table, we have the following observations:(1) MCN is substantially beneficial to $F_{1}$ improvement, which indicates the information flow carried by the proposed document-level graph can enhance the dependencies between entities. (2) BERT encoder and ranking loss contributes 2\% $F_{1}$ and 0.76\% $F_{1}$ improvement respectively. (3) By linking mentions with Wikidata, our model benefits from its accurate relation annotations. Overall, our BERT-MCN model outperforms sequence-based models, attention-based models and parse tree-based models.

\subsection{Model Analysis}
In this subsection, we demonstrate the effectiveness of each component using the development set of DocRED.

\begin{table}
\caption{ Edge ablations on dev set (\%). We remove the ranking loss of BERT-MCN for fair comparison}
\label{edge type}
\centering
\begin{tabular}{l|c}
\hline \textbf{Model} &\textbf{F1} \\ \hline
BERT-MCN &57.00 \\
\hline
- \textit{MM} &54.00\\
- \textit{ME} &54.72\\
- \textit{MS} &55.76\\
- \textit{ES} &56.36\\
\hline
GCNN & 51.52\\
BiLSTM-ADJ &52.00\\
BERT-ADJ &55.30\\
\hline
\end{tabular}
\end{table}

\subsubsection{Aggressive Mention Linking Matters}
Compared with GCNN which limits the exchange of information to neighbouring sentences or inner-entity mentions, our model is able to exchange information between heterogeneous mention pairs regardless of distance. To investigate the usefulness of aggressive mention linking strategy, we restrict $MM$ edge in 2.2.2 to only link mentions belonging to neighboring sentences(the same sentence include) or the same entities. As shown in Table \ref{edge type}, the GCNN-similar implementation of BERT-MCN, BERT-ADJ,  achieves 55.30\% F1 score in development set, which indicates the inference of cross-sentence mention pairs is the key of document-level relation extraction. Implementation of BiLSTM-ADJ achieves 52.00\% F1 score in development set.

Additionally, we remove each type of edge in the constructed graph one by one to examine their effectiveness. As shown in Table \ref{edge type}, removal of MM and ME edges significantly degrades the model's performance while MS and ES edges do not significantly affect the performance. The result provides another evidence that the edges about mentions are the critical points of document-level relation extraction.

\begin{figure}
    \centering
    \includegraphics[width=0.8\textwidth]{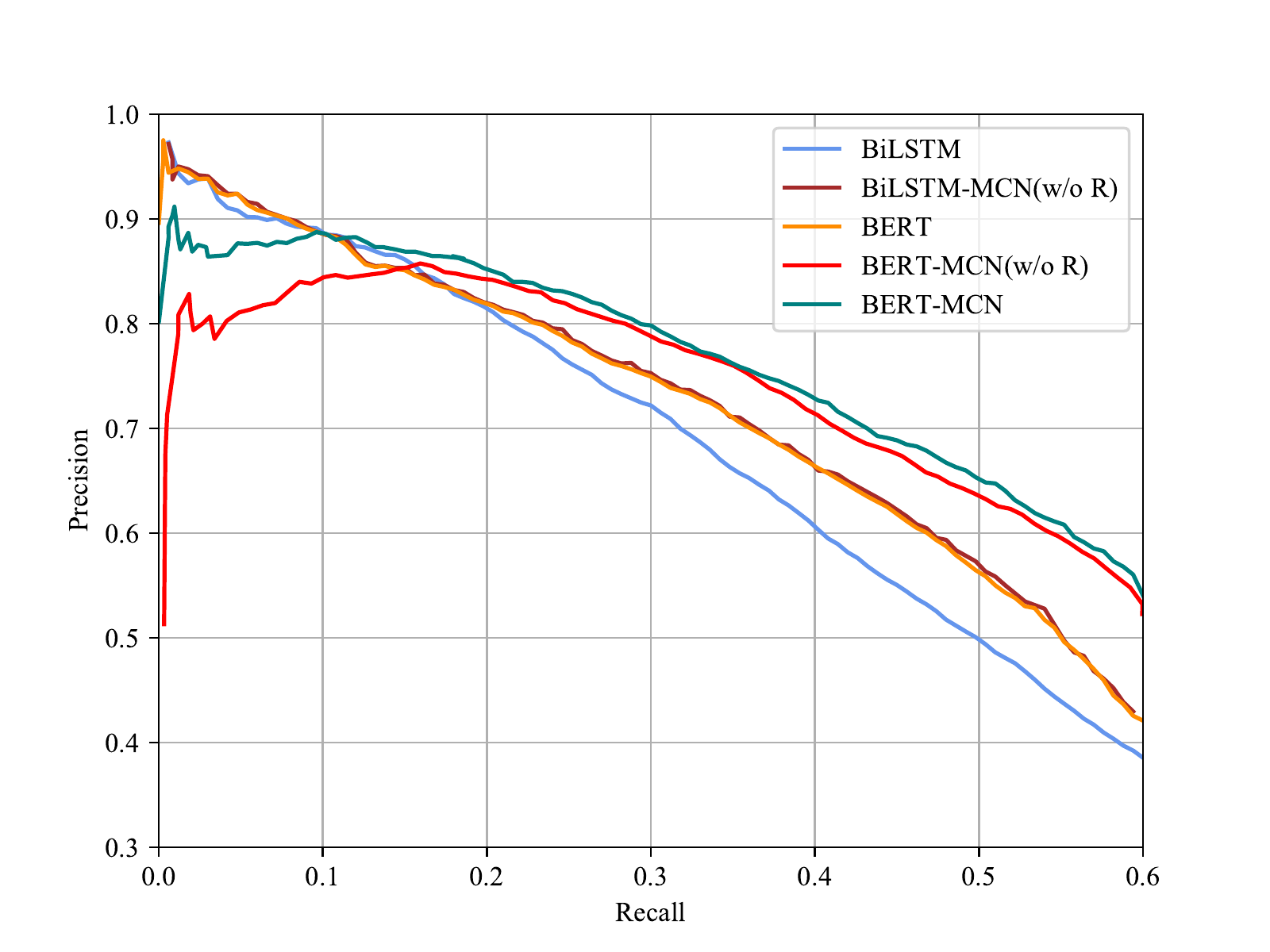}
    \caption{Aggregate precision/recall curves of different models on DocRED}
    \label{PR_curve}
\end{figure}

\subsubsection{Ranking Loss and NA Class}
We present PR curves of BiLSTM, BiLSTM-MCN(w/o ranking loss), BERT , BERT-MCN(w/o ranking loss), BERT-MCN models in Fig \ref{PR_curve}. As shown in the figure, a peculiar sharp decline occurs in BERT-MCN(w/o R) model in the low recall area, which sharply hurts the performance but does not happen in other models. In order to find out what happens in this sharp decline, we analyse the incorrect samples from top 10\% recall area. Surprisingly, we discover most samples are indeed correct or partially alluded by the text but not included in the annotations. During the annotation process of DocRED, because of the large number of potential entity pairs in the DocRED, \cite{yao-etal-2019-docred} first generate triplet candidates from RE models and distant supervision based on entity linking, then ask human annotators to label these candidates. This process inevitably ignores some instances which traditional models are not good at, randomizing the distribution of DocRED. By replacing BCE loss with our proposed ranking loss, The PR curve of BERT-MCN is much smoother by circumventing predicting polluted NA instances in the training set. This problem also indicates the performance of our proposed model could be underestimated. We further relax restrictions if the entity pairs predict the highest none NA relations which are negative but listed in Wikidata. As Table \ref{F1 score} shows, our model gets additional 1.2\% $F_{1}$ improvement.

\subsubsection{Intra- and Inter-sentence Performance }
\begin{figure}
    \centering
    \includegraphics[width=0.75\textwidth]{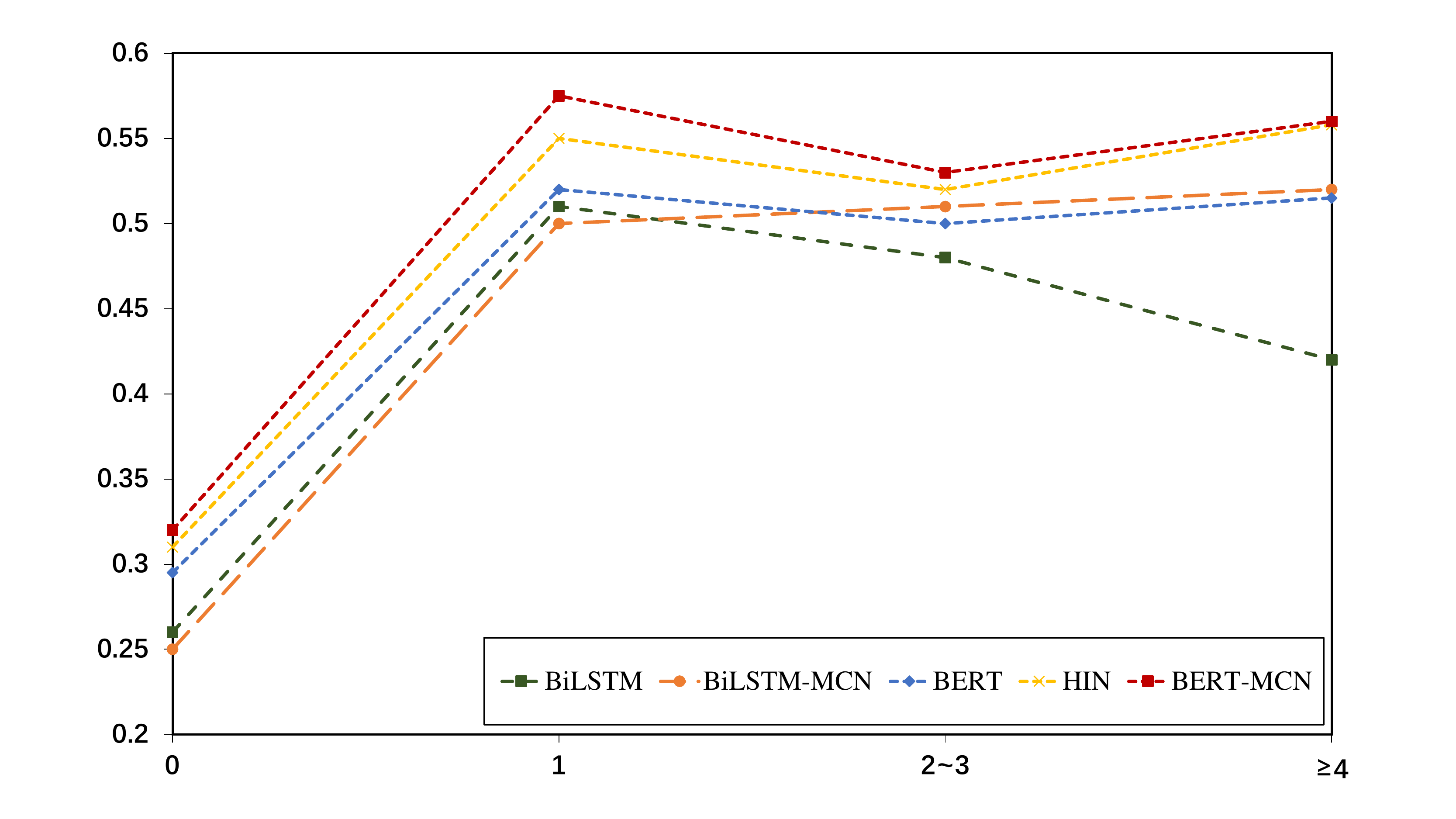}
    \caption{Models' recall of relational facts with different number of supporting evidences.}
    \label{evidence_number}
\end{figure}

 It is obvious that more supporting evidences indicate the model should consider more information from other sentences. According to the number of supporting evidences of gold relations in dev set, we divide them into $d=0/1/2-3$ (0/1/2 or 3 supporting evidences) and $d\geq4$ (more than 4 supporting evidences), and then analyse the recall on relational facts in Fig \ref{evidence_number}$\footnote{We omit BERT-LSR for the lack of source code}$. Apparently, MCN greatly boost the accuracy especially when the supporting evidences are large, which illustrates the MCN's ability to synthesize the information between multiple sentences.  

\subsection{Case Study}
\begin{figure}
    \centering
    \includegraphics[width=1\textwidth]{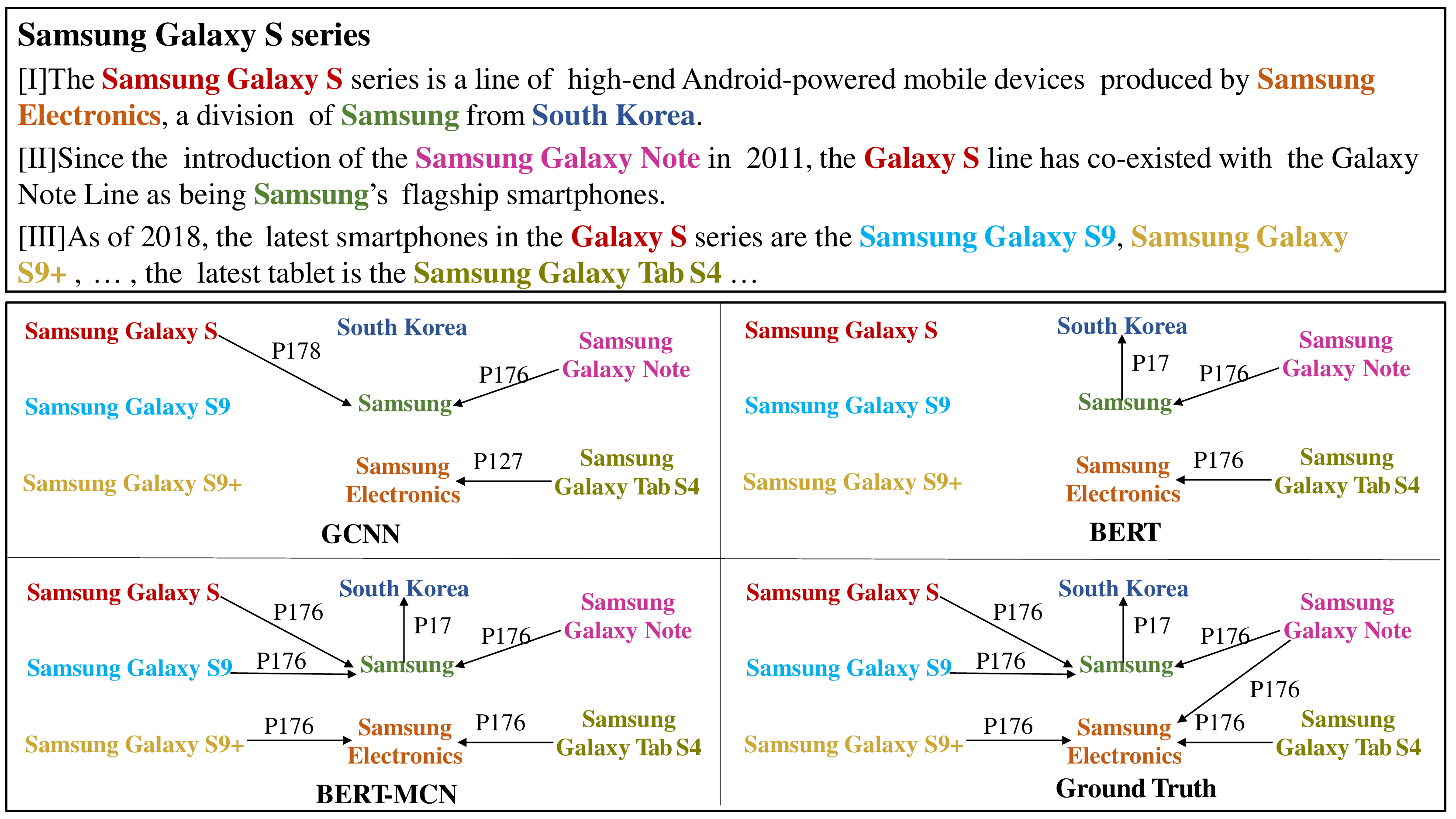}
    \caption{Case study of ``Samsung Galaxy S series'' of DocRED. We visualize the predicted relational facts by different models in the lower column. ``P17'' refers to ``country''. ``P127'' refers to ``owned by''. ``P176'' refers to ``manufacturer'' and ``P178'' refers to ``developer''. Some relations and words are omitted for brevity.}
    \label{fig:case_study}
\end{figure}

Figure \ref{fig:case_study} presents some relational facts predicted by our BERT-MCN model and two baselines. 

\paragraph{Commonsense reasoning:} ``Samsung'' and ``South Korea'' has a relation ``country'' which can be identified by the word ``from'' in sentence (\uppercase\expandafter{\romannumeral1}). GCNN model(as well as BiLSTM model) fails to recognize this pattern possibly because ``country'' is more often related with another preposition ``in''. However, models with BERT layer successfully predicts this relation. Additionally, BERT model successfully predicts 2 of 3 ``manufacturer'' relational facts connected with ``Samsung Electronics''. After consulting wikipedia, we find the text ``produced by Samsung Electronics'' appears in both pages of correct mentions while the incorrect ``Samsung Galaxy S9+'' does not appear in the original text. We argue BERT may encode commonsense knowledge during pre-training process while BiLSTM encoder relies more on pattern recognition. 

\paragraph{Logical reasoning:} BERT-MCN model successfully predicts every ``manufacturer'' relational facts connected with ``Samsung'' while others do not. To identify this relation, the model first needs to identity ``Samsung Electronics'' produces ``Samsung Galaxy S'' and ``Samsung Electronics'' belongs to ``Samsung'' in sentence (\uppercase\expandafter{\romannumeral1}), then identify these products such as ``Samsung Galaxy S9'' belong to ``Galaxy S'' from sentence (\uppercase\expandafter{\romannumeral3}). We argue our MCN structure collects compositional information from various intermediate mention pairs, so that it can discover complicated higher-order inter-sentence relationships. 

\section{Related Work}
Traditional approaches in RE focus on sentence-level relation extraction, using CNN \cite{zeng-etal-2014-relation} or RNN \cite{ebrahimi-dou-2015-chain} to encode sentences as relation representation. Later works add attention mechanism\cite{lin-etal-2016-neural,zeng-etal-2015-distant}, reinforcement learning \cite{feng2018reinforcement,zeng2018large}, generative adversarial network \cite{qin-etal-2018-dsgan} and capsule network \cite{zhang2019multi} to deal with distant supervision\cite{riedel2010modeling} setting. The existing approaches which deal with document-level relation extraction mainly focus on medical relations.  \cite{quirk-poon-2017-distant} propose the first approach for applying distant supervision to cross-sentence relation extraction. They add an edge between the dependency roots of adjacent sentences and extract features from the graph to copy with inter-sentence relations in GDKD. Later \cite{peng-etal-2017-cross,gupta2019neural} extend this method by introducing Graph LSTMs or Dependency-Based RNN to encode the whole graph. But they only consider up to 3 consecutive sentences. Recently, \cite{jia-etal-2019-document} propose to classify relation from entity-level pairs rather than mention-level pairs. \cite{christopoulou-etal-2019-connecting} construct similar document-level graph, but they limit mention interactions within one sentence and construct relation-dependent edge representations by attention scores between entity nodes. \cite{tang2020hin} leverages translation constrict to capture relation representation between entities and aggregates information across the document by attention scores between entities and sentences. 

Recent GCN methods used in RE most utilize the dependency tree to construct graph. \cite{zhang-etal-2018-graph} use the GCN layer to encode the dependency path for relation extraction and achieves state-of-art performance on TACRED dataset. \cite{guo-etal-2019-attention} extend this method by introducing attention mechanism to extract a more precise dependency graph. \cite{fu-etal-2019-graphrel} first perform GCN operation on dependency tree and then performs a 2nd-phase prediction based on relation-weighted graph to consider interaction between named entities and relations. \cite{sahu-etal-2019-inter} construct document-level graph by linking adjacent roots of parse trees and coreference mentions. \cite{nan-etal-2020-reasoning} dynamically learn a document-level graph through structured attention of shortest dependency trees nodes and Matrix-Tree Theorem.

\section{Conclusion and Future Work}
In this paper, we establish cross-sentence dependencies through document-level fully-connected mention pairs. Moreover, considering our model is sensitive to widespread mislabeled NA instances of document-level relation extraction dataset, we propose an improved version of ranking loss to generalize relational representations for mislabeled NA instances. Experimental results show our model achieves comparable results with previous methods which rely on parse trees or attention mechanism. Our future work aims to design more subtle ways to exchange information between node representations of document-level graphs. 

%
%
%
\bibliographystyle{splncs04}

\end{document}